\documentclass{article}
\usepackage{spconf,amsmath,graphicx}
\usepackage{amsmath,amssymb}
\usepackage{multirow}

\usepackage{pifont}
\usepackage{xcolor}
\newcommand{\cmark}{\ding{51}}
\newcommand{\xmark}{\ding{55}} 

\usepackage{hyperref}
\hypersetup{
  pdfauthor={},
  pdftitle={},
  pdfsubject={},
  pdfkeywords={}
}

\title{Egocentric Whole-Body Human Mesh Recovery with Prior-Guided Learning}
\name{Soyeon Na \qquad Seung Young Noh \qquad Ju Yong Chang\thanks{This work was supported by Institute of Information \& Communications Technology Planning \& Evaluation (IITP) grant funded by the Korea government (MSIT) (No. RS-2023-00219700, Development of FACS-compatible Facial Expression Style Transfer Technology for Digital Human).}}
\address{Dept of ECE, Kwangwoon University, Seoul, Korea\\
{\small \texttt {\{naso06,kelvinnoh,jychang\}@kw.ac.kr}}}

\begin{document}
%
\maketitle
\begin{abstract}
Egocentric human mesh recovery (HMR) from monocular head-mounted cameras is increasingly important for AR/VR applications, but remains challenging due to the lack of reliable ground-truth (GT) annotations based on parametric human body models such as SMPL and SMPL-X for real egocentric images. Existing egocentric HMR methods typically rely on pseudo-GT and focus on body pose estimation, which limits their ability to recover fine-grained whole-body details such as hands and face. We study egocentric whole-body human mesh recovery and propose a prior-guided learning framework that reconstructs whole-body meshes from a single egocentric image. We construct more accurate optimization-based pseudo-GT aligned with 3D joint supervision, and leverage multiple priors by adapting an exocentric HMR foundation model together with a diffusion-based pose prior. A deterministic undistortion module is further adopted to handle fisheye distortions in egocentric images. Experiments across multiple egocentric benchmarks demonstrate improved whole-body reconstruction compared to state-of-the-art methods, and show that our optimization-based pseudo-GT is substantially more accurate than existing regression-based pseudo-GT. To facilitate reproducibility, the code and dataset annotations are publicly available at \url{https://github.com/naso06/EgoSMPLX}.
\end{abstract}
\begin{keywords}
Egocentric vision, whole-body human mesh recovery, prior-guided learning
\end{keywords}

\begin{figure*}[t]
  \centering
  \includegraphics[width=\linewidth]{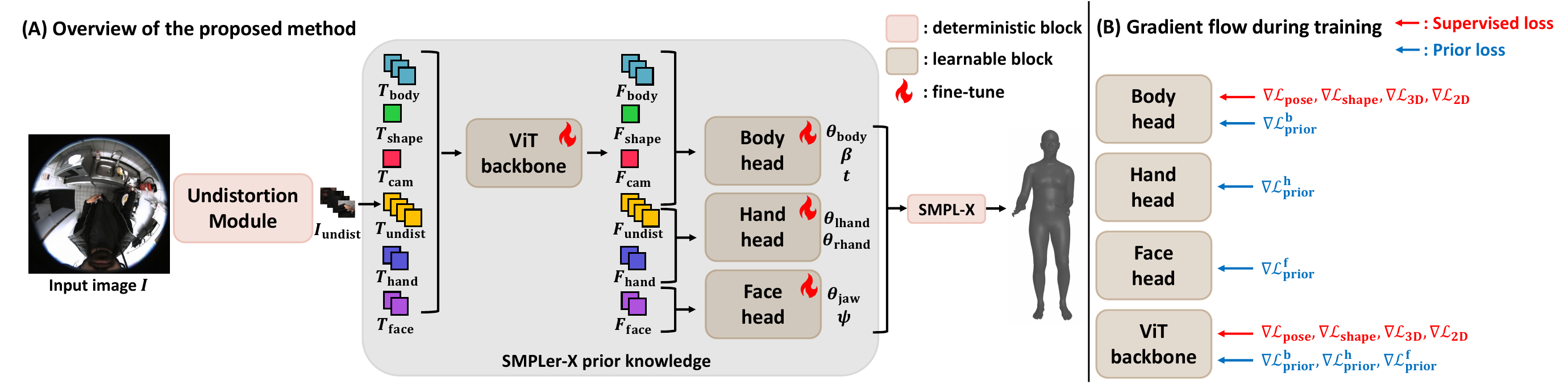}
  \vspace{-7mm}
  \caption{\textbf{(A): Overview of the proposed method.} Given an input egocentric image $I$, an undistortion module outputs undistorted patches $I_{\mathrm{undist}}$, which are fed into a ViT backbone initialized from SMPLer-X and fine-tuned using our pseudo-GT. The predicted parameters are converted into a 3D mesh via the SMPL-X body model. \textbf{(B): Gradient flow during training.} Since our pseudo-GT provides body-only annotations, \textcolor{red}{supervised losses} backpropagate gradients only to the body head and the shared ViT backbone. We additionally apply a diffusion-based plausible pose prior over the full SMPL-X pose, and backpropagate \textcolor{blue}{prior loss} gradients to all pose heads and the shared backbone without requiring explicit GT annotations.}
  \vspace{-3mm}
  \label{fig:method}
\end{figure*}

\section{Introduction}
\label{sec:intro}

Egocentric human mesh recovery (HMR) aims to reconstruct the 3D pose and shape of a head-mounted device (HMD) wearer from monocular egocentric images captured by a HMD-mounted camera. With the increasing adoption of metaverse and AR/VR technologies, accurately reconstructing humans from an egocentric (first-person) viewpoint has become increasingly important, leading to growing research interest in egocentric HMR. Meanwhile, exocentric HMR takes third-person images as input and has been extensively studied over the years, making it a well-established research area with diverse applications. Most HMR methods in both egocentric and exocentric settings rely on parametric human body models such as SMPL~\cite{SMPL:2015} or SMPL-X~\cite{SMPL-X:2019} to represent the 3D body and estimate their pose and shape parameters. While SMPL effectively models full-body pose and shape, it does not capture fine-grained hand and facial details, such as hand poses and facial expressions. In contrast, SMPL-X provides a unified whole-body representation that jointly models the body, hands, and face.

Since exocentric HMR has been actively studied over the years, several datasets provide accurate SMPL(-X) ground-truth (GT) annotations for real images, enabled by motion capture systems or multi-view camera setups. In contrast, egocentric HMR is relatively recent, and datasets with reliable SMPL(-X) GT annotations for real egocentric images remain limited. As a result, many existing egocentric HMR methods~\cite{liu2023egohmr,shen2025fish2mesh} generate and use SMPL pseudo-GT during training. For example, \cite{liu2023egohmr} leverages a dataset that provides paired egocentric and exocentric views and adopts SMPL parameters estimated on exocentric images by an off-the-shelf exocentric HMR model~\cite{kocabas2021pare} as pseudo-GT. However, such pseudo-GT is constructed without explicitly enforcing consistency with available 3D pose annotations, and thus can be inaccurate for heavily occluded or difficult poses. These inaccuracies can destabilize training and ultimately degrade performance.

Most existing egocentric HMR approaches therefore rely on SMPL pseudo-GT and primarily focus on estimating body pose parameters. Because SMPL pose parameters are defined only over body joints, hands are often reconstructed in a default pose, making accurate hand pose estimation challenging. This limitation is particularly restrictive for gesture-based interaction and immersive applications in egocentric settings, where fine-grained whole-body reconstruction, including hands and face, is essential.

To address these challenges, we study egocentric whole-body human mesh recovery and propose \emph{a prior-guided learning framework that effectively compensates for limited egocentric supervision}. We first construct more accurate optimization-based pseudo-GT SMPL annotations aligned with GT 3D joint supervision, providing stronger body pose supervision than prior regression-based pseudo annotations~\cite{liu2023egohmr}. To further recover body parts without direct annotations, such as hands and face, we leverage multiple priors. Specifically, we fine-tune an exocentric HMR foundation model, SMPLer-X~\cite{cai2023smpler}, and incorporate a diffusion-based pose prior, DPoser-X~\cite{lu2025dposer}, to provide plausible supervision beyond available GT.

Despite their effectiveness, directly adapting exocentric priors to egocentric images remains challenging due to strong fisheye distortions commonly observed in egocentric cameras. Learning such distortions from limited egocentric data can harm the generalizable priors acquired from large-scale exocentric training~\cite{wang2022cotta}. To mitigate this issue, we adopt a deterministic undistortion module~\cite{wang2024egowholemocap} to preprocess egocentric inputs before training, enabling distortion-aware learning while preserving the robustness of the transferred priors.

\textbf{Contributions.}
This paper makes the following contributions.
(1) We introduce the first egocentric whole-body human mesh recovery approach from monocular egocentric images.
(2) We build more accurate optimization-based pseudo-GT annotations aligned with 3D joint supervision.
(3) We leverage multiple priors by fine-tuning an exocentric HMR foundation model and incorporating a diffusion-based pose prior (DPoser-X) to compensate for limited egocentric supervision.
(4) We adopt a deterministic undistortion module to handle fisheye distortions in egocentric images.

\section{Proposed Method}
\label{sec:proposed_method}

\subsection{Overview}
\label{ssec:overview}

We propose an SMPL-X-based framework for egocentric whole-body human mesh recovery. As illustrated in Fig.~\ref{fig:method}, the proposed pipeline consists of a fisheye undistortion module, a vision transformer (ViT)~\cite{dosovitskiy2021vit} encoder, and three regression heads that predict SMPL-X parameters for the body, hands, and face from a monocular egocentric image.

To exploit whole-body priors learned from large-scale exocentric data, we adapt the exocentric HMR foundation model SMPLer-X~\cite{cai2023smpler} to the egocentric setting. Since egocentric images often exhibit severe fisheye distortion, we rectify the input using a deterministic undistortion module and feed undistorted image tokens to the ViT encoder. This preprocessing improves feature compatibility with the exocentric pretraining domain and facilitates stable adaptation.

During fine-tuning, we address the scarcity of reliable SMPL-X annotations in egocentric datasets using a hybrid supervision strategy. Specifically, we supervise the body head with optimization-based pseudo-GT body annotations, while regularizing the hand and face heads using a diffusion-based pose prior loss (DPoser-X)~\cite{lu2025dposer} in the absence of direct supervision. This design enables effective egocentric adaptation while preserving the useful whole-body priors learned from exocentric data.

\subsection{Encoder}
\label{ssec:encoder}
Egocentric images often exhibit strong fisheye distortion, which introduces a domain mismatch when directly applying a ViT pretrained on exocentric data and can degrade feature representations. To reduce this domain gap, we first rectify the input using a deterministic undistortion module and feed the resulting undistorted patches to the encoder.

\textbf{Undistortion module.}
Given an input image $I$, the undistortion module produces undistorted image patches $I_{\mathrm{undist}}$. These patches are converted into image tokens via patch embedding and positional encoding, resulting in undistorted image tokens $T^{i}_{\mathrm{undist}}$. We follow the procedure in~\cite{wang2024egowholemocap} to compute the undistorted patches.

\textbf{ViT backbone.}
The ViT backbone takes the undistorted image tokens $T^{i}_{\mathrm{undist}}$ together with a set of learnable task tokens $T^{i}_{\mathrm{task}}=\{T^{i}_k\}_{k\in\mathcal{K}}$ and outputs $(T^{o}_{\mathrm{undist}},\,T^{o}_{\mathrm{task}})$, where $T^{o}_{\mathrm{task}}=\{T^{o}_k\}_{k\in\mathcal{K}}$. Here, the task token set $\mathcal{K}=\{\mathrm{body},\mathrm{shape},\mathrm{cam},\mathrm{hand},\mathrm{face}\}$ follows the SMPLer-X architecture and provides task-specific conditioning for downstream parameter regression. The output undistorted image tokens $T^{o}_{\mathrm{undist}}$ are shared across all heads, while the task tokens serve as latent representations that guide the prediction of different SMPL-X parameter groups.

\subsection{Decoder}
\label{ssec:decoder}

We employ three task-specific regression heads for the body, hands, and face. Each head takes the shared undistorted image tokens $T^{o}_{\mathrm{undist}}$ together with the corresponding task tokens $T^{o}_{\mathrm{task}}$ and regresses the associated SMPL-X parameters.

\textbf{Body head.}
The body head estimates 3D body joints as an intermediate representation and regresses SMPL-X body pose parameters $\boldsymbol{\theta}_{\mathrm{body}}$ and shape parameters $\boldsymbol{\beta}$. From the undistorted image tokens $T^{o}_{\mathrm{undist}}$, we predict a 3D body heatmap $H_{\mathrm{body}}$ and obtain 3D joints via soft-argmax. The estimated 3D joints are combined with the body task token $T^{o}_{\mathrm{body}}$ to regress $\boldsymbol{\theta}_{\mathrm{body}}$, while the shape task token $T^{o}_{\mathrm{shape}}$ estimates $\boldsymbol{\beta}$. We also predict a camera-relative root translation $\mathbf{t}$ to support geometric projection and supervision.

\textbf{Hand head.}
To predict hand pose parameters $\{\boldsymbol{\theta}_{\mathrm{lhand}},$ $\boldsymbol{\theta}_{\mathrm{rhand}}\}$, we first localize the hands, since they occupy small regions in input images. We predict left and right hand bounding boxes using $T^{o}_{\mathrm{undist}}$ and the body heatmap $H_{\mathrm{body}}$. Using the predicted bounding boxes, we crop hand-aligned features from $T^{o}_{\mathrm{undist}}$ and predict a 3D hand heatmap $H_{\mathrm{hand}}$. We obtain 3D hand joints via soft-argmax and combine them with the hand task token $T^{o}_{\mathrm{hand}}$ to regress hand pose parameters.

\textbf{Face head.}
The face head regresses facial expression parameters $\boldsymbol{\psi}$ and jaw pose $\boldsymbol{\theta}_{\mathrm{jaw}}$ using the face task token $T^{o}_{\mathrm{face}}$. Further architectural and implementation details for the ViT backbone and heads follow SMPLer-X~\cite{cai2023smpler}.

\subsection{Loss Function}
\label{ssec:loss}

We train our method with a composite objective that combines SMPL-X parameter supervision, joint-level geometric supervision, and a diffusion-based whole-body pose prior:
\begin{align}
\mathcal{L}_{\text{total}}
&= \lambda_{\text{pose}}\mathcal{L}_{\text{pose}}
 + \lambda_{\text{shape}}\mathcal{L}_{\text{shape}}
 + \lambda_{3D}\mathcal{L}_{3D} \nonumber\\
&\quad + \lambda_{2D}\mathcal{L}_{2D}
 + \lambda_{\text{prior}}\mathcal{L}_{\text{prior}},
\label{eq:loss_total}
\end{align}
where $\lambda_{\text{pose}}, \lambda_{\text{shape}}, \lambda_{3D}, \lambda_{2D}, \lambda_{\text{prior}}$ are scalar weights. 

Let $(\boldsymbol{\theta}_{\mathrm{body}}, \boldsymbol{\beta})$ denote the predicted SMPL-X body pose and shape parameters, and $(\boldsymbol{\theta}^{*}_{\mathrm{body}}, \boldsymbol{\beta}^{*})$ the pseudo-GT targets obtained via optimization. Since pseudo-GT is available only for body pose and shape, we apply $\ell_2$ losses:
\begin{equation}
\mathcal{L}_{\text{pose}} = \left\| \boldsymbol{\theta}_{\mathrm{body}} - \boldsymbol{\theta}^{*}_{\mathrm{body}} \right\|_2^2,
\qquad
\mathcal{L}_{\text{shape}} = \left\| \boldsymbol{\beta} - \boldsymbol{\beta}^{*} \right\|_2^2 .
\label{eq:loss_pose_shape}
\end{equation}

Let $\mathbf{J}_{3D}$ be the predicted 3D joints regressed from the predicted SMPL-X parameters and $\mathbf{J}^{*}_{3D}$ the GT 3D joints. We supervise 3D joints directly with
\begin{equation}
\mathcal{L}_{3D} = \left\| \mathbf{J}_{3D} - \mathbf{J}^{*}_{3D} \right\|_2^2.
\label{eq:loss_3d}
\end{equation}

For 2D supervision, we impose a reprojection loss by projecting translated 3D joints with the fisheye projection $\pi(\cdot)$ under Scaramuzza’s camera model~\cite{scaramuzza2006toolbox}:
\begin{equation}
\mathcal{L}_{2D} = \left\| \mathbf{J}_{2D} - \mathbf{J}^{*}_{2D} \right\|_2^2,
\qquad
\mathbf{J}_{2D} = \pi\!\left(\mathbf{J}_{3D} + \mathbf{t}\right),
\label{eq:loss_2d}
\end{equation}
where $\mathbf{t}$ denotes the predicted camera-relative root translation in the egocentric camera coordinate system. Although the network takes undistorted images, reprojection is performed in the original fisheye camera model for geometric consistency.

To encourage anatomically plausible whole-body poses and mitigate missing hand and face supervision, we impose a diffusion-based pose prior on the full SMPL-X pose. Let $\boldsymbol{\theta} = \{\boldsymbol{\theta}_{\mathrm{body}}, \boldsymbol{\theta}_{\mathrm{lhand}}, \boldsymbol{\theta}_{\mathrm{rhand}}, \boldsymbol{\theta}_{\mathrm{jaw}}\}$ denote the predicted whole-body pose parameters. The prior loss is defined as
\begin{equation}
\mathcal{L}_{\text{prior}} = \left\| \boldsymbol{\theta} - \boldsymbol{\theta}_{0}(\tau) \right\|_2^2 ,
\label{eq:loss_prior}
\end{equation}
where $\boldsymbol{\theta}_{0}(\tau)$ is a one-step denoised estimate from perturbing $\boldsymbol{\theta}$ at noise level $\tau$ and denoising it with a pretrained diffusion model~\cite{lu2025dposer}. This prior provides plausible supervision for the full body, including hands and face, without GT annotations.

\subsection{Pseudo-GT Optimization}
\label{sec:optimization}

To obtain reliable pseudo-GT SMPL annotations for egocentric images, we employ an optimization-based fitting procedure inspired by SMPLify-X~\cite{SMPL-X:2019}. Given GT 3D body joints $\mathbf{J}^{*}_{3D}$ associated with an egocentric image, we estimate the SMPL body pose $\boldsymbol{\theta}_{\mathrm{body}}$ and shape $\boldsymbol{\beta}$ by minimizing:
\begin{align}
E(\boldsymbol{\theta}_{\mathrm{body}}, \boldsymbol{\beta})
&= E_{J_{3D}} + \lambda_{\theta} E_{\theta} + \lambda_{\beta} E_{\beta},
\label{eq:pseudogt_total}
\end{align}
where $\lambda_{\theta}$ and $\lambda_{\beta}$ are loss weights. The regularization term $E_{\theta}$ encourages anatomically plausible body poses, while $E_{\beta}$ constrains the estimated body shape to remain close to the human shape distribution learned from training data.

The data term $E_{J_{3D}}$ enforces consistency between the predicted 3D joints obtained from the SMPL parameters and the GT 3D body joints:
\begin{equation}
E_{J_{3D}} = \sum_{i \in \mathcal{J}} \omega\!\left( \left\| \mathbf{J}_{3D,i} - \mathbf{J}^{*}_{3D,i} \right\|_2 \right),
\end{equation}
where $\mathcal{J}$ denotes the set of annotated body joints, and $\mathbf{J}_{3D,i}$ and $\mathbf{J}^{*}_{3D,i}$ represent the predicted and GT 3D joint locations for joint $i$, respectively. $\omega(\cdot)$ is the robust Geman–McClure function~\cite{GemanMcClure1987}, which reduces the influence of outlier joints.

Compared to the original SMPLify-X formulation, which relies on a 2D reprojection term based on detected 2D keypoints, we replace the 2D fitting term with a direct 3D joint fitting term using the available GT 3D joints $\mathbf{J}^{*}_{3D}$. This modification leads to more accurate and stable pseudo-GT annotations for egocentric data. Since most existing egocentric datasets provide annotations only for body joints, our fitting procedure focuses on obtaining reliable pseudo-GT for the body pose and shape, rather than full-body details such as hands and face. The resulting optimized parameters $(\boldsymbol{\theta}_{\mathrm{body}}, \boldsymbol{\beta})$ are used as pseudo-GT annotations for training.

\section{Experimental Results}

\subsection{Implementation Details}

We resize all input images $I$ to $256\times256$. Each image is patchified into $16\times16$ non-overlapping patches, resulting in $256$ image tokens, which are fed into the undistortion module. To align with the input resolution of the ViT backbone, we crop the undistorted patches by removing two columns from each side, yielding $12\times16=192$ undistorted patches. For training, we set the loss weights to $\lambda_{\text{pose}}=10$, $\lambda_{\text{shape}}=10^{-2}$, $\lambda_{3D}=10^{2}$, and $\lambda_{2D}=1$. The diffusion prior weight $\lambda_{\text{prior}}$ is decayed from $10^{-1}$ to $10^{-2}$ using a cosine schedule to stabilize training. For the pseudo-GT optimization objective, we set $\lambda_{\theta}=10^{3}$ and $\lambda_{\beta}=10^{2}$. We initialize the proposed method with pretrained SMPLer-X weights and fine-tune it on the proposed dataset for $5$ epochs. Optimization is performed using Adam with a batch size of $8$ and a cosine learning-rate schedule decaying from $10^{-5}$ to $5\times10^{-7}$. All experiments are conducted on $4$ NVIDIA RTX 4090 GPUs.

\subsection{Datasets}

We evaluate our method on both real-world and synthetic egocentric datasets. EgoPW~\cite{wang2022egopw} is a real-world egocentric dataset captured using a head-mounted fisheye camera synchronized with an external third-person camera. The head-mounted camera provides an egocentric view, while the external camera captures the same scene from a third-person perspective. The EgoPW training set contains 133K frames from 6 subjects across 8 indoor daily-life scenes (e.g., kitchen and living room), and the test set consists of 114K frames from 3 unseen subjects. SceneEgo~\cite{wang2023scenego} is a real-world egocentric dataset focusing on diverse human--scene interactions. Its test set contains 12K frames from 2 subjects performing activities such as sitting on a chair, reading a newspaper, and using a computer. We use SceneEgo exclusively for cross-dataset evaluation to assess generalization to unseen scenes and interaction patterns. EgoWholeBody~\cite{wang2024egowholemocap} is a large-scale synthetic egocentric dataset with full-body 3D pose and SMPL-X annotations. It contains approximately 700K rendered frames from 14 RenderPeople identities. For evaluation, we use the training split and uniformly subsample one frame every 100 frames, resulting in approximately 7K frames, which allows efficient evaluation while covering diverse poses and viewpoints. We train our model on the EgoPW training set and evaluate it on the EgoPW test set, the SceneEgo test set, and EgoWholeBody.

\subsection{Evaluation Metrics}

We evaluate the proposed method using Procrustes-aligned mean per-joint position error (PA-MPJPE) and mean per-vertex position error (PA-MPVPE), both measured in mm. In egocentric HMR, many existing methods do not explicitly estimate camera-relative translation, and evaluation based on raw MPJPE or MPVPE can therefore be dominated by global misalignment between predicted and GT meshes. To focus on the accuracy of reconstructed pose and shape rather than global alignment, we apply Procrustes alignment before computing joint- and vertex-level errors.

\begin{table}[t]
    \centering
    \caption{Quantitative comparison with state-of-the-art methods on the EgoPW, SceneEgo, and EgoWholeBody datasets. The best results are shown in bold.}
    \vspace{2mm}
    \label{tab:quantitative_result}
    \setlength{\tabcolsep}{3pt}
    \renewcommand{\arraystretch}{1.0}
    \scriptsize
    {\fontsize{7.5}{10}\selectfont
    \begin{tabular}{lllcc}
    \hline
    Dataset & Body region & Method & PA-MPJPE ($\downarrow$) & PA-MPVPE ($\downarrow$) \\
    \hline
    \multirow{3}{*}{EgoPW} & \multirow{3}{*}{Body-only} & EgoHMR & 88.97  & 58.81 \\
    & & Fish2Mesh & 101.54 & 61.94 \\
    & & Ours & \textbf{87.28} & \textbf{53.71} \\
    \hline
    \multirow{3}{*}{SceneEgo} & \multirow{3}{*}{Body-only} & EgoHMR & 111.07 & 83.13 \\
    & & Fish2Mesh & 171.97 & 106.47 \\
    & & Ours & \textbf{50.86} & \textbf{38.69} \\
    \hline
    \multirow{3}{*}{EgoWholeBody} & \multirow{3}{*}{Whole-body} & EgoHMR & 167.22 & 114.98 \\
    & & Fish2Mesh & 194.93 & 125.58 \\
    & & Ours & \textbf{57.75} & \textbf{43.17} \\
    \hline
    \multirow{3}{*}{EgoWholeBody} & \multirow{3}{*}{Hand-only} & EgoHMR & 18.60 & 19.04 \\
    & & Fish2Mesh & 18.60 & 19.02 \\
    & & Ours & \textbf{17.07} & \textbf{17.38} \\
    \hline
    \end{tabular}
    \vspace{-4mm}
}
\end{table}

\subsection{Comparison with State-of-the-art Methods}
\label{ssec:quant_results}

\textbf{Quantitative results.}
We compare our method with state-of-the-art egocentric HMR approaches on EgoPW, SceneEgo, and EgoWholeBody. For fair comparison, all methods are trained only on the EgoPW training set and evaluated on the remaining benchmarks without additional training. While EgoHMR~\cite{liu2023egohmr} and Fish2Mesh~\cite{shen2025fish2mesh} predict SMPL parameters, our method estimates SMPL-X parameters. To enable evaluation under a unified body model representation, we convert the SMPL outputs of EgoHMR and Fish2Mesh to SMPL-X format using an off-the-shelf SMPL-to-SMPL-X conversion tool~\cite{sarandi2024smplfitter}. This conversion is applied only during evaluation and does not involve learning or additional supervision. Similarly, for EgoPW and SceneEgo, whose pseudo-GT meshes follow the SMPL topology, we convert the GT meshes to SMPL-X for evaluation. EgoWholeBody provides SMPL-X GT by default and therefore requires no conversion.

Table~\ref{tab:quantitative_result} reports quantitative results on EgoPW, SceneEgo, and EgoWholeBody. Since EgoPW and SceneEgo provide annotations for body joints only, we evaluate the body-only setting on these datasets. On EgoWholeBody, which includes whole-body joint annotations, we report results for both whole-body and hand-only settings. Across all benchmarks, our method consistently achieves the lowest error under all evaluation settings. Notably, in the whole-body setting, our approach significantly outperforms prior methods. In the hand-only evaluation, existing SMPL-based methods tend to predict less articulated hand poses, whereas our method produces more detailed and expressive hand reconstructions.

\begin{figure}[t]
  \centering
  \includegraphics[width=\linewidth]{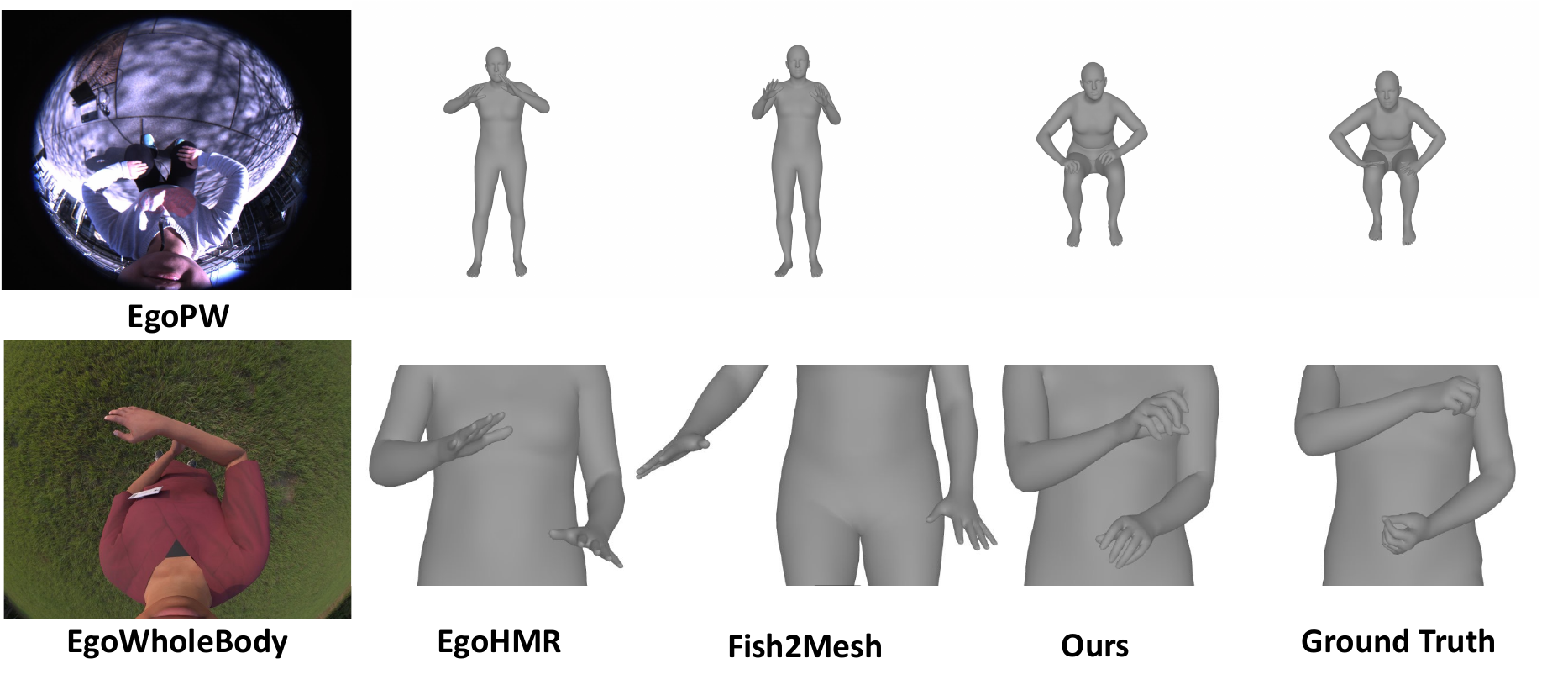}
  \vspace{-9mm}
  \caption{Qualitative comparison on EgoPW (whole-body, top) and EgoWholeBody (hand close-ups, bottom).} 
  \label{fig:qual_compare}
\end{figure}

\textbf{Qualitative results.}
Fig.~\ref{fig:qual_compare} qualitatively demonstrates the effectiveness of our method on EgoPW and EgoWholeBody. Existing methods often produce reconstructions that are inconsistent with the input egocentric images, such as predicting standing poses even when the subject is seated, as shown in the top row of Fig.~\ref{fig:qual_compare}. In contrast, our method generates image-consistent mesh reconstructions that better reflect the observed body configurations. Moreover, while prior SMPL-based approaches tend to produce default or weakly articulated hand poses due to limited hand representation, our SMPL-X-based method predicts more detailed hand poses, resulting in more realistic whole-body reconstructions, as shown in the bottom row of Fig.~\ref{fig:qual_compare}.

\begin{table}[t]
\centering
\caption{Ablation study on the SceneEgo test dataset. ``Prior'' denotes leveraging prior knowledge from SMPLer-X~\cite{cai2023smpler}, $I_{\text{undist}}$ indicates the use of undistorted images via the undistortion module, and $\mathcal{L}_{\text{prior}}$ denotes the diffusion-based pose prior loss~\cite{lu2025dposer}. The last row corresponds to the full model.}
\vspace{2mm}
\label{tab:ablation}
\setlength{\tabcolsep}{4pt}
\renewcommand{\arraystretch}{1.05}
{\fontsize{9}{11}\selectfont
\begin{tabular}{c|ccc|cc}
\hline
Setting & Prior & $I_\text{undist}$ & $\mathcal{L}_{\text{prior}}$ & PA-MPJPE ($\downarrow$) & PA-MPVPE ($\downarrow$) \\
\hline
(i) &  &  &  & 108.29 & 76.82 \\
(ii) & \cmark &  &  & 72.10 & 51.87 \\
(iii) & \cmark & \cmark &  & 58.81 & 43.55 \\
\hline
(iv) & \cmark & \cmark & \cmark & \textbf{50.86} & \textbf{38.69} \\
\hline
\end{tabular}
}
\vspace{-2mm}
\end{table}

\subsection{Ablation Study}
\label{ssec:ablation}

\textbf{Ablation on model components.}
We conduct an ablation study on the SceneEgo test dataset to quantify the contribution of each component in the proposed method. Table~\ref{tab:ablation} reports PA-MPJPE and PA-MPVPE for the following variants: 
(i) a baseline trained from scratch using our pseudo-GT annotations without exocentric pretraining, 
(ii) a model fine-tuned from the exocentric foundation model SMPLer-X~\cite{cai2023smpler} to leverage its prior knowledge, 
(iii) additionally applying the deterministic fisheye undistortion module~\cite{wang2024egowholemocap} and using undistorted input images $I_{\text{undist}}$, and 
(iv) the full model with the diffusion-based pose prior loss $\mathcal{L}_{\text{prior}}$~\cite{lu2025dposer}.

Compared to the scratch baseline, fine-tuning from SMPLer-X yields a clear improvement, indicating that priors learned from large-scale third-person data transfer effectively to the egocentric domain. Incorporating the undistortion module further reduces the error by mitigating strong fisheye distortions and facilitating stable adaptation of exocentric priors. Finally, adding the diffusion-based pose prior achieves the best performance, demonstrating that the learned prior effectively regularizes training under limited supervision and suppresses anatomically implausible predictions.

\begin{table}[t]
\centering
\caption{Comparison of pseudo-GT quality on the EgoPW training set. PA-MPJPE is computed between pseudo-GT and GT 3D joints. ``Aligned to GT'' indicates whether pseudo-GT is aligned with GT 3D joints.}
\vspace{2mm}
\label{tab:pseudogt_quality}
\setlength{\tabcolsep}{9pt}
\renewcommand{\arraystretch}{1.05}
{\fontsize{9}{11}\selectfont
\begin{tabular}{l|c|c}
\hline
Pseudo-GT method & Aligned to GT & PA-MPJPE ($\downarrow$) \\
\hline
Regression via~\cite{kocabas2021pare} & \xmark & 350.78 \\
Optimization (Ours) & \cmark & \textbf{77.44} \\
\hline
\end{tabular}
}
\vspace{-2mm}
\end{table}

\textbf{Pseudo-GT quality.}
Previous work often derives pseudo-GT from exocentric images using different pipelines. One strategy applies a regression-based exocentric HMR model; for instance, EgoHMR~\cite{liu2023egohmr} adopts PARE~\cite{kocabas2021pare}. Another line of work~\cite{shen2025fish2mesh} estimates pseudo-GT via human--scene optimization~\cite{liu20214d_human}. These optimization-based methods reconstruct the scene from a monocular video captured with camera motion using COLMAP~\cite{schonberger2016structure} and optimize human body model parameters under human--scene constraints. However, EgoPW uses a fixed exocentric camera, offering insufficient parallax for reliable COLMAP reconstruction, making direct comparison difficult. Notably, existing pseudo-GT generation pipelines, regardless of whether they are regression-based or optimization-based, still depend on exocentric images and typically do not exploit available GT 3D joints.

We evaluate pseudo-GT quality on EgoPW in Table~\ref{tab:pseudogt_quality}. We extract 3D joints from each pseudo-GT SMPL mesh and compute PA-MPJPE against GT 3D joints. Regression-based pseudo-GT yields large errors, whereas our optimization-based pseudo-GT reduces PA-MPJPE by approximately 78\%, providing more reliable pose supervision.

\section{Conclusion}
\label{sec:conclusion}

We studied egocentric whole-body human mesh recovery from monocular egocentric images and presented a prior-guided framework that enables robust reconstruction under limited egocentric supervision. By combining optimization-based pseudo-GT aligned with 3D joint annotations, exocentric whole-body priors, and a diffusion-based pose prior, our approach effectively recovers articulated body, hand, and face poses from challenging egocentric inputs. Incorporating temporal information from egocentric video could improve stability and reduce ambiguity in severely occluded poses. Extending the framework to jointly reason about human--scene interactions may enhance physical plausibility in everyday environments. Finally, scaling the approach to more diverse real-world egocentric datasets with richer annotations would help advance egocentric whole-body understanding toward practical AR/VR applications.

\twocolumn[
\begin{center}
{\Large\bfseries
Supplementary Material: Egocentric Whole-Body Human Mesh Recovery with Prior-Guided Learning\par}
\vspace{12mm}
\end{center}
]

\renewcommand{\thefigure}{S\arabic{figure}}
\renewcommand{\thetable}{S\arabic{table}}
\renewcommand{\thesection}{S\arabic{section}}
\renewcommand{\theequation}{S\arabic{equation}}

\setcounter{figure}{0}
\setcounter{table}{0}
\setcounter{section}{0}
\setcounter{equation}{0}

\section{Overview}

In this supplementary material, we provide additional details and analyses that were omitted from the main paper due to space constraints. Section~\ref{sec:add_impl} describes the implementation details of the undistortion module, including the fisheye camera model, undistorted patch generation, and image tokenization process. Section~\ref{sec:add_analysis} presents additional analyses of computational cost and hyperparameter stability.

\section{Additional Implementation Details}
\label{sec:add_impl}

\subsection{Undistortion Module}
\label{sec:undistortion_module}

\textbf{Fisheye camera model.}
Following previous work~\cite{wang2024egowholemocap}, we adopt the fisheye camera model proposed by Scaramuzza~\cite{scaramuzza2006toolbox}. The fisheye projection function $\mathcal{P}$ maps a 3D point $\mathbf{p}=(x,y,z)$ to fisheye image coordinates $(u,v)$ as follows:
\begin{equation}
\label{eq_s1:projection_pre}
    \rho = \arctan\left(\frac{z}{\sqrt{x^2 + y^2}}\right),
    \qquad
    f(\rho) = \sum_{q=0}^{Q} k_q \rho^q,
\end{equation}
\begin{equation}
\label{eq_s2:projection_define}
    \mathcal{P}(x,y,z) =
    \begin{bmatrix} u \\ v \end{bmatrix} =
    f(\rho)\frac{\begin{bmatrix} x \\ y \end{bmatrix}}{\sqrt{x^2 + y^2}},
\end{equation}
where $k_q$ denotes the polynomial coefficients obtained through camera calibration.

Given a 2D point $(u,v)$ on the fisheye image, the corresponding 3D point along the viewing ray is obtained using the fisheye reprojection function $\mathcal{P}^{-1}$:
\begin{equation}
\label{eq_s3:reprojection_pre}
    \rho' = \sqrt{u^2 + v^2}, \qquad
    f'(\rho') = \sum_{q=0}^{Q'} k'_q (\rho')^q,
\end{equation}
\begin{equation}
\label{eq_s4:reprojection_define}
    \mathcal{P}^{-1}(u,v,a) =
    \begin{bmatrix} x \\ y \\ z \end{bmatrix}
    =
    a \cdot
    \frac{
    \begin{bmatrix} u \\ v \\ f'(\rho') \end{bmatrix}}
    {\sqrt{u^2 + v^2 + (f'(\rho'))^2}},
\end{equation}
where $k'_q$ denotes the coefficients of another polynomial obtained through camera calibration, and $a$ represents the distance between the camera center and the 3D point $(x, y, z)$.

\begin{figure}[t]
  \centering
  \includegraphics[width=\linewidth]{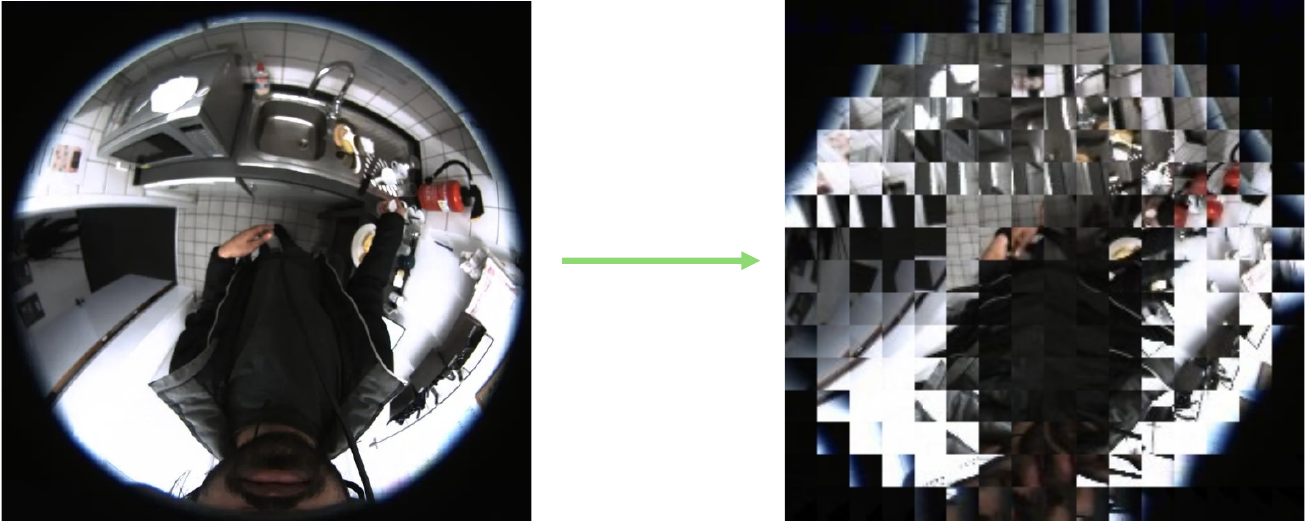}
  \caption{\textbf{Visualization of undistorted patch generation.} Square patches are defined in the undistorted space and projected onto the fisheye image, where pixel values are obtained via bilinear sampling.}
  \label{fig:undistortion_patch}
\end{figure}

\textbf{Undistorted patch generation.}
Using the fisheye projection and reprojection functions $\mathcal{P}$ and $\mathcal{P}^{-1}$ defined above, the undistortion module generates undistorted patches through a deterministic geometric procedure. Instead of directly extracting square patches from distorted fisheye images, patches are defined in the undistorted space and projected onto the fisheye image to obtain pixel values via bilinear sampling. Fig.~\ref{fig:undistortion_patch} illustrates the input fisheye image and the resulting undistorted patches.

Given an input image $I$ of size $H \times W$, we first sample $N \times N$ patch center points $(u_i,v_j)$ in the fisheye image coordinate space:
\begin{equation}
    \mathbf{c}_{ij} = (u_i, v_j) =
    \left(
    \frac{W}{N}\left(i + \frac{1}{2}\right),
    \frac{H}{N}\left(j + \frac{1}{2}\right)
    \right),
\end{equation}
where $i,j \in \{0,\dots,N-1\}$.

Each patch center point is mapped onto the unit sphere using the fisheye reprojection function $\mathcal{P}^{-1}$ defined in Eq.~\ref{eq_s4:reprojection_define}:
\begin{equation}
\label{eq_s6:center_point}
    \mathbf{p}_{ij}^{c} = (x_{ij}^{c}, y_{ij}^{c}, z_{ij}^{c}) = \mathcal{P}^{-1}(u_i, v_j, 1).
\end{equation}

To determine the orientation of the local grid on the tangent plane, we sample a neighboring image point:
\begin{equation}
    \mathbf{p}_{ij}^{u} = (x_{ij}^{u}, y_{ij}^{u}, z_{ij}^{u})
    = \mathcal{P}^{-1}(u_i + d, v_j, 1),
\end{equation}
where $d$ denotes a horizontal offset in the fisheye image coordinate space, as in~\cite{wang2024egowholemocap}.

The intersection point $\mathbf{p}_{ij}^{x}$ between the ray from the origin passing through $\mathbf{p}_{ij}^{u}$ and the tangent plane passing through $\mathbf{p}_{ij}^{c}$ is computed as:
\begin{equation}
\label{eq_s8:intersection_point}
    \mathbf{p}_{ij}^{x} 
    = \frac{\langle \mathbf{p}_{ij}^{c}, \mathbf{p}_{ij}^{c} \rangle}
    {\langle \mathbf{p}_{ij}^{u}, \mathbf{p}_{ij}^{c} \rangle}
    \mathbf{p}_{ij}^{u},
\end{equation}
where $\langle \cdot, \cdot \rangle$ denotes the inner product.

Using the center and intersection points defined above, we define the local coordinate axes $\mathbf{v}_{ij}^{x}$, $\mathbf{v}_{ij}^{z}$, and $\mathbf{v}_{ij}^{y}$ as follows:
\begin{equation}
    \mathbf{v}_{ij}^{x} = \frac{\mathbf{p}_{ij}^{x} - \mathbf{p}_{ij}^{c}} {\left\| \mathbf{p}_{ij}^{x} - \mathbf{p}_{ij}^{c} \right\|},
    \qquad
    \mathbf{v}_{ij}^{z} = \frac{\mathbf{p}_{ij}^{c}}
    {\left\| \mathbf{p}_{ij}^{c} \right\|},
\end{equation}
\begin{equation}
    \mathbf{v}_{ij}^{y} = \frac{\mathbf{v}_{ij}^{z} \times \mathbf{v}_{ij}^{x}} {\left\| \mathbf{v}_{ij}^{z} \times \mathbf{v}_{ij}^{x} \right\|},
\end{equation}
where $\|\cdot\|$ denotes the Euclidean norm.

Using the local coordinate axes $\mathbf{v}_{ij}^{x}$ and $\mathbf{v}_{ij}^{y}$, we define an $M \times M$ sampling grid within an $l \times l$ square on the tangent plane:
\begin{equation}
    \mathbf{p}_{ij}^{mn} = \mathbf{p}_{ij}^{c} + \frac{l}{M}\left(\tilde{m}\mathbf{v}_{ij}^{x} + \tilde{n}\mathbf{v}_{ij}^{y}\right),
\end{equation}
where $\tilde{m}=m-\frac{M-1}{2}$, $\tilde{n}=n-\frac{M-1}{2}$, and $m,n \in \{0,\dots,M-1\}$. Here, $M$ is the patch sampling resolution, and $l$ controls the scale of the square grid on the tangent plane.

Each sampled point on the tangent plane is projected back onto the fisheye image using the fisheye projection function $\mathcal{P}$ defined in Eq.~\ref{eq_s2:projection_define}:
\begin{equation}
    \mathbf{c}_{ij}^{mn} = \mathcal{P}(\mathbf{p}_{ij}^{mn}).
\end{equation}

Finally, the undistorted patch $I_{ij}^{\mathrm{undist}}$ is obtained by bilinear sampling from the input image $I$ at the projected coordinates $\mathbf{c}_{ij}^{mn}$:
\begin{equation}
    I_{ij}^{\mathrm{undist}}(m,n) = \mathrm{Bilinear}(I, \mathbf{c}_{ij}^{mn}).
\end{equation}

This process generates one undistorted patch per center point. Repeating the procedure for all $N \times N$ locations yields a set of undistorted patches. Boundary patches are discarded to match the input resolution of the ViT backbone. Following~\cite{wang2024egowholemocap}, we use $N=16$, $M=16$, $d=8$, and $l=0.2$ for input images of size $H=W=256$. The resulting patches are then converted into image tokens as described in Sec.~\ref{sec:image_tokenization}. Additional implementation details of the undistortion patch generation procedure can be found in EgoWholeMocap~\cite{wang2024egowholemocap}.

\subsection{Image Tokenization}
\label{sec:image_tokenization}

Given the undistorted patches generated in Sec.~\ref{sec:undistortion_module}, we follow the standard ViT tokenization process~\cite{dosovitskiy2021vit}. Each patch is converted into an image token through patch embedding, and positional encoding is added to preserve spatial information.

This process does not introduce a new tokenization scheme. Instead, the main difference from conventional ViT input processing lies in the preceding patch generation step, where patches are defined in the undistorted space and sampled from the fisheye image.

\section{Additional Analysis}
\label{sec:add_analysis}

\subsection{Computational Cost Analysis}

We analyze the computational cost of our method during inference and compare it with previous egocentric human mesh recovery methods~\cite{liu2023egohmr,shen2025fish2mesh}. Table~\ref{tab:complexity} reports GFLOPs, the number of parameters, model size, inference time, and FPS measured during inference.

\begin{table}[t]
    \centering
    \caption{Comparison of computational cost with previous methods during inference.}
    \vspace{2mm}
    \label{tab:complexity}
    \resizebox{\columnwidth}{!}{
    \begin{tabular}{lccccc}
    \hline
    Method & GFLOPs & \#Params & Model Size (MB) & Time (ms) & FPS \\
    \hline
    EgoHMR~\cite{liu2023egohmr} & 13.51 & 107.58M & 410.96 & 39.40 & 25.35 \\
    Fish2Mesh~\cite{shen2025fish2mesh} & 4.74 & 7.50M & 48.12 & 3.90 & 332.36 \\
    Ours & 191.87 & 663.26M & 2713.08 & 45.79 & 21.84 \\
    \hline
    \end{tabular}
    }
\end{table}

Our method requires more GFLOPs and parameters than the compared methods, primarily due to the high-capacity ViT backbone~\cite{cai2023smpler}. Despite this, the practical runtime remains comparable to EgoHMR~\cite{liu2023egohmr}, achieving 45.79 ms per frame and 21.84 FPS. Among the model components used during inference, the ViT backbone accounts for most of the parameters (632.39M), while the undistortion module is deterministic and parameter-free. The regression heads and neck modules contain 4.89M and 25.98M parameters, respectively. The diffusion-based prior, DPoser-X~\cite{lu2025dposer}, contains 25.81M parameters but is used only during training. Therefore, it is excluded from the test-time parameter count, model size, GFLOPs, inference time, and FPS reported in Table~\ref{tab:complexity}. Overall, the increased test-time model size mainly reflects the backbone capacity rather than overhead from the diffusion prior. All runtime measurements are conducted on a single NVIDIA RTX 3090 GPU with a batch size of 1.

\subsection{Hyperparameter Stability Analysis}

We analyze the stability of the proposed method under different hyperparameter configurations, including the learning rate schedule and loss weights. Specifically, we compare two learning rate schedules: $1\times10^{-4}$--$5\times10^{-6}$ and $5\times10^{-5}$--$1\times10^{-6}$. Both settings exhibit stable convergence behavior and yield comparable reconstruction accuracy. On the EgoPW dataset, the corresponding PA-MPJPE values are 66.25 and 64.67, respectively.

These results demonstrate that the proposed method is robust to moderate variations in hyperparameter settings. This suggests that the performance gain is not merely the result of careful hyperparameter tuning, but rather reflects the robustness of the proposed training objective.

\vfill\pagebreak



\bibliographystyle{IEEEbib}
\bibliography{refs}

\end{document}